%% file: top.tex
\documentclass[10pt,twocolumn,letterpaper]{article}

\usepackage[utf8]{inputenc}
\usepackage{cvpr}
\usepackage{times}
\usepackage{graphicx}
\usepackage{amsmath,amsfonts,amssymb,amsthm}

\usepackage{color}
\usepackage{caption}
\usepackage{subcaption}
\usepackage{xspace}
\usepackage{paralist}
\usepackage{array,booktabs,calc}
\usepackage{placeins}
\usepackage{cite}

\usepackage[pagebackref=true,breaklinks=true,letterpaper=true,colorlinks,bookmarks=false]{hyperref}

\DeclareMathOperator{\KL}{KL}

\input{shortcuts}

\cvprfinalcopy 

\ifcvprfinal\fi

\begin{document}

\renewcommand*{\thefootnote}{\fnsymbol{footnote}}

\title{Occupancy Networks: Learning 3D Reconstruction in Function Space}

\author{Lars Mescheder$^{1}$ \quad Michael Oechsle$^{1,2}$ \quad Michael Niemeyer$^{1}$ \quad Sebastian Nowozin$^{3}\footnotemark[2]$ \quad Andreas Geiger$^{1}$\\
	$^1$Autonomous Vision Group, MPI for Intelligent Systems and University of Tübingen \\
	$^2$ETAS GmbH, Stuttgart \\
	$^3$Google AI Berlin \\
	{\tt\small \{firstname.lastname\}@tue.mpg.de \qquad nowozin@gmail.com}	
}

\maketitle
\thispagestyle{empty}

\begin{abstract}
\input{abstract}
\end{abstract}

\begin{figure}[t!]
\centering
\begin{subfigure}{0.25\linewidth}
  \centering
  \includegraphics[width=0.9\linewidth]{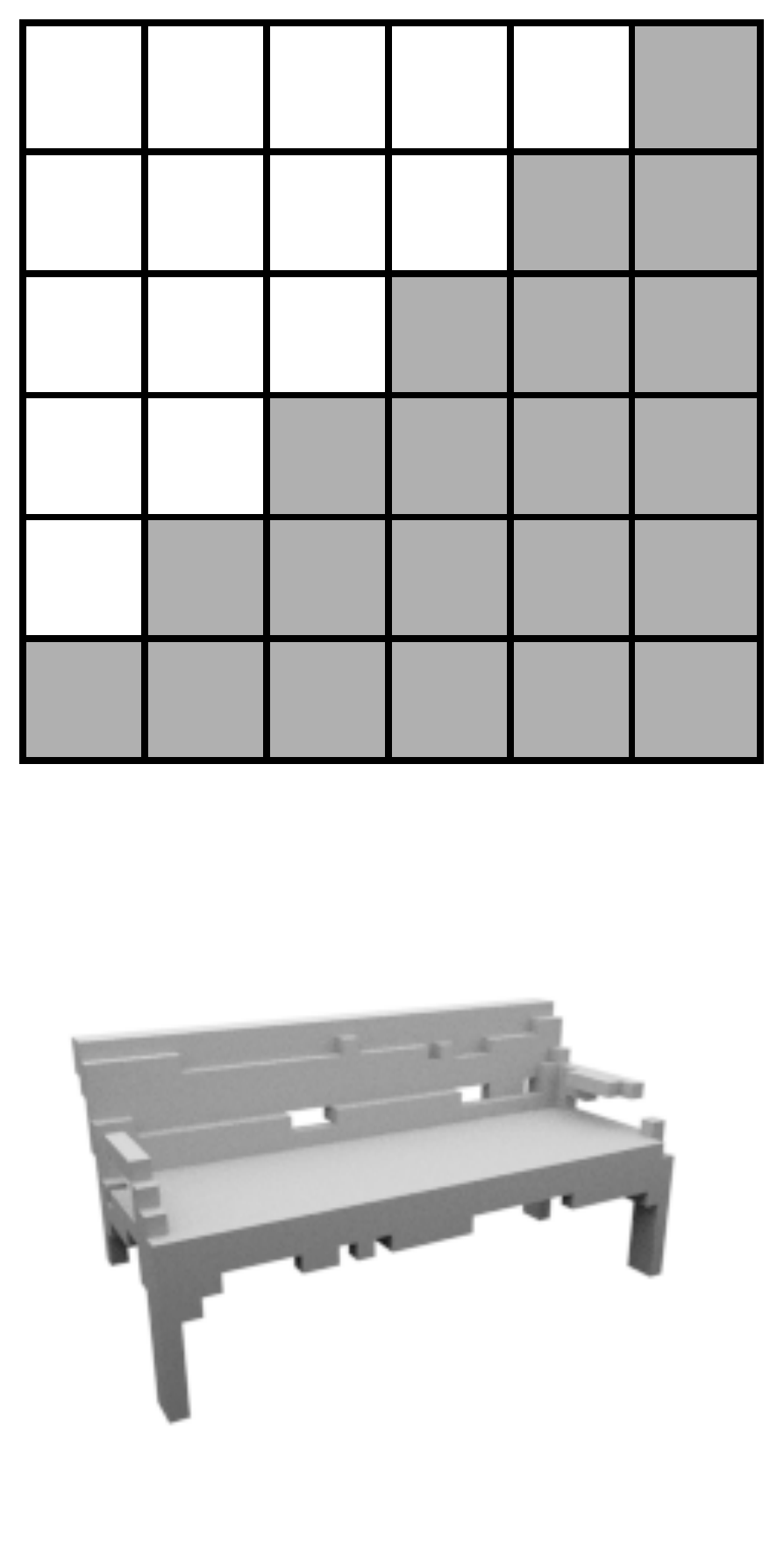}
  \caption{Voxel}\label{fig:teaser_voxel}
\end{subfigure}%
\begin{subfigure}{0.25\linewidth}
  \centering
  \includegraphics[width=0.9\linewidth]{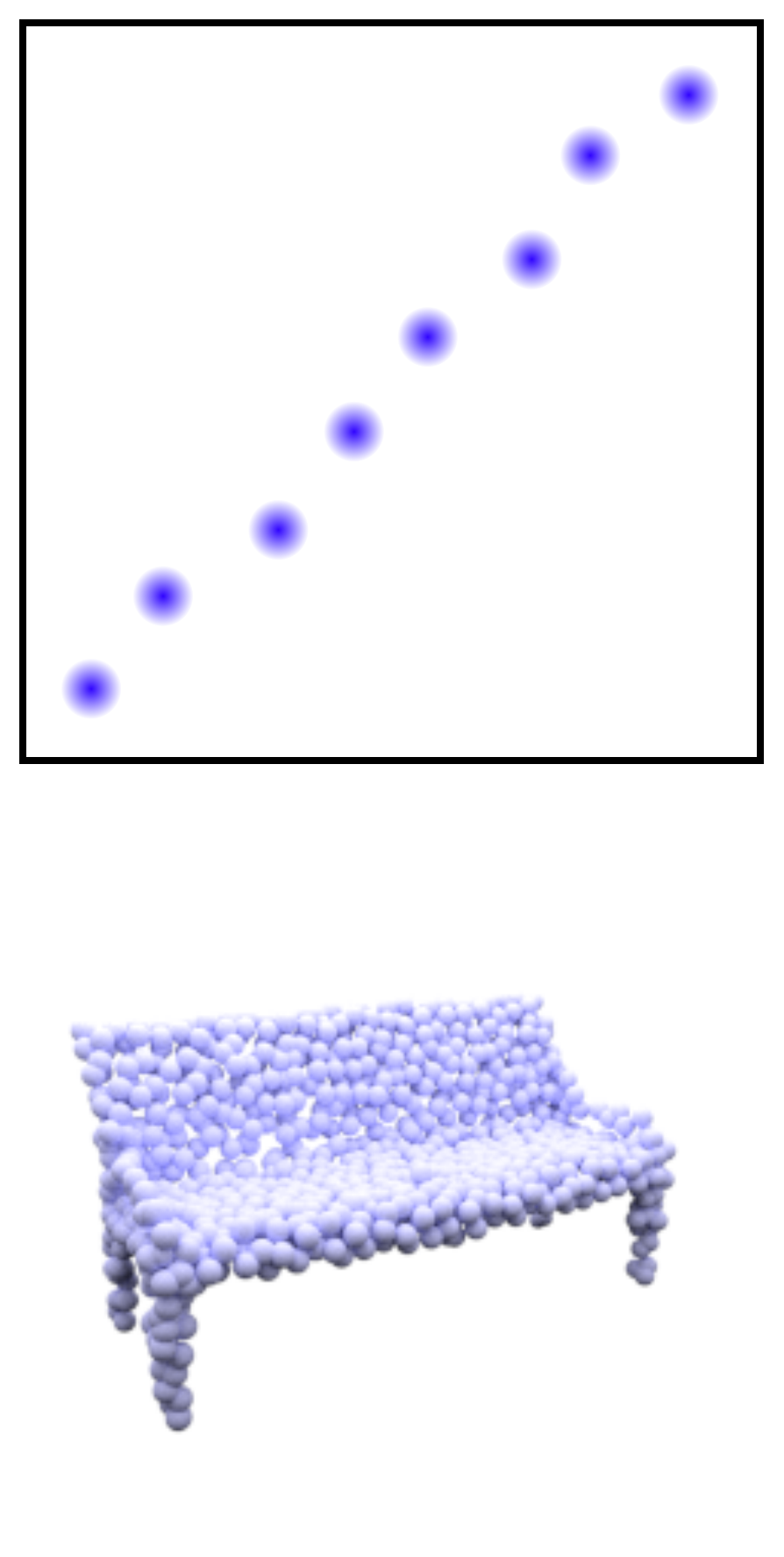}
  \caption{Point}\label{fig:teaser_point}
\end{subfigure}%
\begin{subfigure}{0.25\linewidth}
  \centering
  \includegraphics[width=0.9\linewidth]{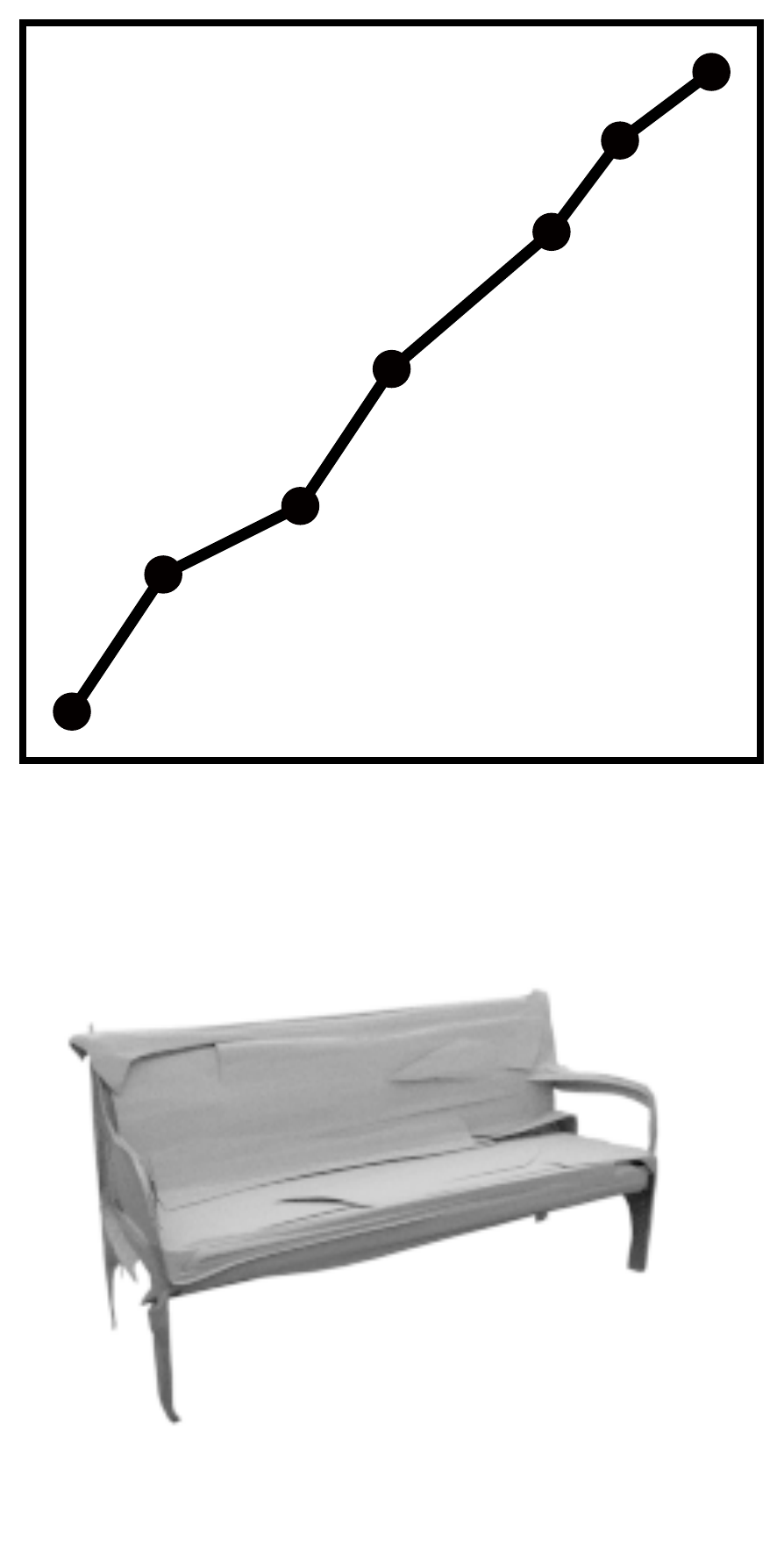}
  \caption{Mesh}\label{fig:teaser_mesh}
\end{subfigure}%
\begin{subfigure}{0.25\linewidth}
  \centering
  \includegraphics[width=0.9\linewidth]{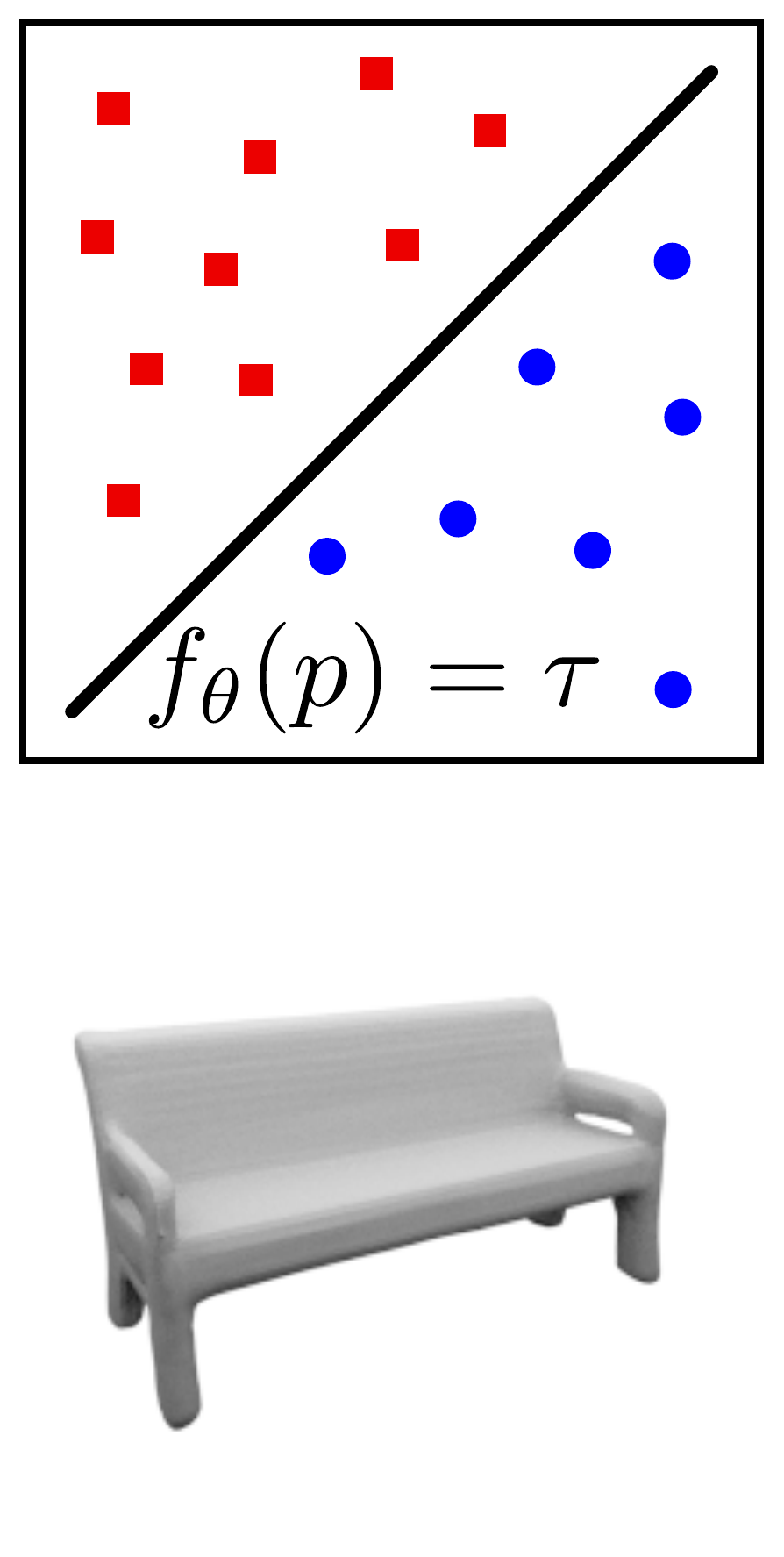}
  \caption{Ours}\label{fig:teaser_ours}
\end{subfigure}%
\caption{
\textbf{Overview:}
Existing 3D representations discretize the output space differently:  (\subref{fig:teaser_voxel}) spatially in voxel representations,  (\subref{fig:teaser_point}) in terms of predicted points, and  (\subref{fig:teaser_mesh}) in terms of vertices for mesh representations.  In contrast, (\subref{fig:teaser_ours}) we propose to consider the continuous decision boundary of a classifier $f_\theta$ (\eg, a deep neural network) as a 3D surface which allows to extract 3D meshes at any resolution.
}
\label{fig:teaser}
\vspace{-0.5cm}
\end{figure}

\footnotetext[2]{Part of this work was done while at MSR Cambridge.}
\renewcommand*{\thefootnote}{\arabic{footnote}}
 \setcounter{footnote}{0}

\input{parts/intro}
\input{parts/related}
\input{parts/method}
\input{parts/experiments}
\input{parts/conclusion}

\section*{Acknowledgements}
This work was supported by the Intel Network on Intelligent Systems and by Microsoft Research through its PhD Scholarship Programme.

\FloatBarrier

{\small
  \bibliographystyle{ieee}
  \bibliography{bib/bibliography_long,bib/bibliography_fixed,bib/bibliography_custom}
}

\end{document}

%% file: shortcuts.tex
\newcommand{\nR}{\mathbb{R}}

\newcommand{\cB}{\mathcal{B}}

\newcommand{\cL}{\mathcal{L}}

\newcommand{\cX}{\mathcal{X}}

\newcommand{\figref}[1]{Fig.~\ref{#1}}

\makeatletter
\DeclareRobustCommand\onedot{\futurelet\@let@token\@onedot}
\def\@onedot{\ifx\@let@token.\else.\null\fi\xspace}
\def\eg{e.g\onedot} 
\def\ie{i.e\onedot} 
 
\def\etc{etc\onedot}

\def\wrt{wrt\onedot}

\def\etal{et~al\onedot}

\makeatother

\newcommand{\boldparagraph}[1]{\vspace{0.2cm}\noindent{\bf #1:} }

\definecolor{darkgreen}{rgb}{0,0.7,0}


%% file: abstract.tex
With the advent of deep neural networks, learning-based approaches for 3D~reconstruction have gained popularity. However, unlike for images, in 3D there is no canonical representation which is both computationally and memory efficient yet allows for representing high-resolution geometry of arbitrary topology. Many of the state-of-the-art learning-based 3D~reconstruction approaches can hence only represent very coarse 3D geometry or are limited to a restricted domain. In this paper, we propose Occupancy Networks, a new representation for learning-based 3D~reconstruction methods. Occupancy networks implicitly represent the 3D surface as the continuous decision boundary of a deep neural network classifier. In contrast to existing approaches, our representation encodes a description of the 3D output at infinite resolution without excessive memory footprint. We validate that our representation can efficiently encode 3D structure and can be inferred from various kinds of input. Our experiments demonstrate competitive results, both qualitatively and quantitatively, for the challenging tasks of 3D reconstruction from single images, noisy point clouds and coarse discrete voxel grids. We believe that occupancy networks will become a useful tool in a wide variety of learning-based 3D tasks.

%% file: parts/intro.tex
\section{Introduction}
Recently, learning-based approaches for 3D reconstruction have gained popularity
~\cite{Wu2015CVPR,Rezende2016NIPS,Choy2016ECCV,Wu2016NIPS,Girdhar2016ECCV,Brock2016ARXIV}.
In contrast to traditional multi-view stereo algorithms, 
learned models are able to encode rich prior information about the space of 3D shapes which helps to resolve ambiguities in the input.

While generative models have recently achieved remarkable successes in generating realistic high resolution images
\cite{Karras2018ICLR,Mescheder2018ICML,Wang2018CVPRa},
this success has not yet been replicated in the 3D domain.
In contrast to the 2D domain, the community has not yet agreed on a 3D output representation that is both memory efficient and can be efficiently inferred from data.
Existing representations can be broadly categorized into three categories: voxel-based representations \cite{Ulusoy2015THREEDV,Wu2016NIPS,Brock2016ARXIV,Gadelha2017THREEDV,Rezende2016NIPS,Stutz2018CVPR,Liao2018CVPR}
, point-based representations 
\cite{Achlioptas2018ICML, Fan2017CVPR}
and mesh representations
\cite{Ranjan2018ECCV,Wang2018ECCV,Kanazawa2018ECCV}, see \figref{fig:teaser}.

Voxel representations are a straightforward generalization of pixels to the 3D case.
Unfortunately, however, the memory footprint of voxel representations grows cubically with resolution, hence limiting na\"ive implementations to $32^3$ or $64^3$ voxels.
While it is possible to reduce the memory footprint by using data adaptive representations such as octrees \cite{Tatarchenko2017ICCV,Riegler2017CVPR}, this approach leads to complex implementations and existing data-adaptive algorithms are still limited to relatively small $256^3$ voxel grids.
Point clouds \cite{Achlioptas2018ICML, Fan2017CVPR} and meshes \cite{Ranjan2018ECCV,Wang2018ECCV,Kanazawa2018ECCV} have been introduced as alternative representations for deep learning, using appropriate loss functions.
However, point clouds lack the connectivity structure of the underlying mesh and hence require additional post-processing steps to extract 3D geometry from the model.
Existing mesh representations are typically based on deforming a template mesh and hence do not allow arbitrary topologies.
Moreover, both approaches are limited in the number of points/vertices which can be reliably predicted using a standard feed-forward network.

In this paper\footnote{%
Also see \cite{Chen2019CVPR,Park2019ARXIV,Michalkiewicz2019Arxiv}
for concurrent work that proposes similar ideas.}, we propose a novel approach to 3D-reconstruction based on directly learning the \textit{continuous} 3D occupancy function (\figref{fig:teaser_ours}).
Instead of predicting a voxelized representation at a fixed resolution, we predict the complete occupancy function 
with a neural network $f_\theta$
which can be evaluated at \emph{arbitrary} resolution.
This drastically reduces the memory footprint during training.
At inference time, we extract the mesh from the learned model using a simple multi-resolution isosurface extraction algorithm which trivially parallelizes over 3D locations.

\noindent In summary, our \textbf{contributions} are as follows:
\begin{compactitem}
 \item We introduce a new representation for 3D geometry based on learning a continuous 3D mapping.
 \item We show how this representation can be used for reconstructing 3D geometry from various input types.
 \item We experimentally validate that our approach is able to generate high-quality meshes and demonstrate that it compares favorably to the state-of-the-art.
 \end{compactitem}

%% file: parts/related.tex
\section{Related Work}\label{sec:related-work}

Existing work on learning-based 3D reconstruction can be broadly categorized by the output representation they produce
as either voxel-based, point-based or mesh-based.

\boldparagraph{Voxel Representations}
Due to their simplicity, voxels are the most commonly used representation for discriminative \cite{Maturana2015IROS,Qi2016CVPR,Song2016CVPR} and generative \cite{Stutz2018CVPR,Wu2015CVPR,Rezende2016NIPS,Choy2016ECCV,Wu2016NIPS,Girdhar2016ECCV}
3D tasks. 

Early works have considered the problem of reconstructing 3D geometry from a single image using 3D convolutional neural networks which operate on voxel grids \cite{Wu2015CVPR,Choy2016ECCV,Tulsiani2017CVPR}. Due to memory requirements, however, these approaches were limited to relatively small $32^3$ voxel grids.
While recent works \cite{Wu2017NIPS,Wu2018ECCV,Zhang2018NIPS} have applied 3D convolutional neural networks to resolutions up to $128^3$, this is only possible with shallow architectures and small batch sizes, which leads to slow training.

The problem of reconstructing 3D geometry from multiple input views has been considered in \cite{Ji2017ICCV,Kar2017NIPS,Paschalidou2018CVPR}.
Ji \etal \cite{Ji2017ICCV} and Kar \etal \cite{Kar2017NIPS} encode the camera parameters together with the input images in a 3D voxel representation and apply 3D convolutions to reconstruct 3D scenes from multiple views.
Paschalidou \etal \cite{Paschalidou2018CVPR} introduced an architecture that predicts voxel occupancies from multiple images, exploiting multi-view geometry constraints \cite{Ulusoy2015THREEDV}.

Other works applied voxel representations to learn generative models of 3D shapes. 
Most of these methods are either based
on variational auto-encoders \cite{Kingma2014ICLR,Rezende2014ICML} 
or generative adversarial networks \cite{Goodfellow2014NIPS}.
These two approaches were pursued in  \cite{Brock2016ARXIV,Rezende2016NIPS} and \cite{Wu2016NIPS}, respectively.

Due to the high memory requirements of voxel representations, recent works have proposed to reconstruct 3D objects in a multi-resolution fashion \cite{Tatarchenko2017ICCV,Hane20173DV}. However, the resulting methods are often complicated to implement and require multiple passes over the input to generate the final 3D model. 
Furthermore, they are still limited to comparably small $256^3$ voxel grids.
For achieving sub-voxel precision, several works~\cite{Dai2017CVPRa,Riegler2017THREEDV,Ladicky2017ICCV} have proposed  to predict truncated signed distance fields (TSDF) ~\cite{Curless1996SIGGRAPH} where each point in a 3D grid stores the truncated signed distance to the closest 3D surface point.
However, this representation is usually much harder to learn compared to occupancy representations as the network must reason about distance functions in 3D space instead of merely classifying a voxel as occupied or not.
Moreover, this representation is still limited by the resolution of the underlying 3D grid.

\boldparagraph{Point Representations}
An interesting alternative representation of 3D geometry is given by 3D point clouds
which are widely used both in the robotics and in the computer graphics communities.
Qi \etal~\cite{Qi2017CVPR, Qi2017NIPS} pioneered point clouds as a representation for discriminative deep learning tasks.
They achieved permutation invariance by applying a fully connected neural network to each point independently followed by a global pooling operation.
Fan \etal~\cite{Fan2017CVPR} introduced point clouds as an output representation for 3D reconstruction.
However, unlike other representations, this approach requires additional non-trivial post-processing steps \cite{Bernardini1999VCG,Kazhdan2006SGP,Kazhdan2013SIGGRAPH,Calakli2011CGF} to generate the final 3D mesh.

\boldparagraph{Mesh Representations}
Meshes have first been considered for discriminative 3D classification or segmentation tasks by applying convolutions on the graph spanned by the mesh's vertices and edges \cite{Bronstein2017SPM,Guo2015SIGGRAPH,Wang2017CG}.

More recently, meshes have also been considered as output representation for 3D reconstruction
\cite{Wang2018ECCV,Groueix2018CVPR,Kong2017CVPR,Kanazawa2018CVPR}.
Unfortunately, most of these approaches are prone to generating self-intersecting meshes.
Moreover, they are only able to generate meshes with simple topology  \cite{Wang2018ECCV}, 
require a reference template from the same object class \cite{Kong2017CVPR,Kanazawa2018CVPR,Ranjan2018ECCV} or
cannot guarantee closed surfaces \cite{Groueix2018CVPR}.
Liao \etal \cite{Liao2018CVPR} proposed an end-to-end learnable version of the marching cubes algorithm \cite{Lorensen1987SIGGRAPH}.
However, their approach is still limited by the memory requirements of the underlying 3D grid and hence also restricted to $32^3$ voxel resolution.

In contrast to the aforementioned approaches, our approach leads to high resolution closed surfaces without self-intersections
and does not require template meshes from the same object class as input.
This idea is related to
classical level set \cite{Dervieux1980, Osher1988, Cremers2007} approaches to multi-view 3D reconstruction~\cite{Faugeras1997,Yezzi2003,Goldluecke2004CVPR, Jin2004CVPR, Pons2005CVPR, Kolev2006}. However, instead of solving a differential equation, our approach uses deep learning to obtain a more expressive representation which can be naturally integrated into an end-to-end learning pipeline.

%% file: parts/method.tex
\section{Method}
In this section, we first introduce \emph{Occupancy Networks} as a representation of 3D geometry.
We then describe how we can learn a model that infers this representation from various forms of input
such as point clouds, single images and low-resolution voxel representations.
Lastly, we describe a technique for extracting high-quality 3D meshes from our model at test time.

\subsection{Occupancy Networks}
Ideally, we would like to reason about the occupancy not only at fixed discrete 3D locations (as in voxel respresentations) but at \emph{every} possible 3D point $p \in \nR^3$.
We call the resulting function
\begin{equation}
o: \nR^3 \to \{0, 1\}
\label{eq:occupancy_function}
\end{equation}
the \emph{occupancy function} of the 3D object.
Our key insight is that we can approximate this 3D function with a neural network that assigns to every location $p \in \nR^3$ an occupancy probability between $0$ and $1$.
Note that this network is equivalent to a neural network for binary classification, except that we are interested in the decision boundary which implicitly represents the object's surface.

When using such a network for 3D reconstruction of an object based on observations of that object (\eg, image, point cloud, \etc), we must condition it on the input.
Fortunately, we can make use of the following simple functional equivalence:
a function that takes an observation $x\in\cX$ as input and has a function from $p \in \nR^3$ to $\nR$ as output can be equivalently described by a function that takes a pair $(p,x) \in \nR^3\times \cX$ as input and outputs a real number.
The latter representation can be simply parameterized by a neural network $f_\theta$ that takes a pair $(p,x)$ as input and outputs a real number which represents the probability of occupancy:
\begin{equation}
  f_\theta: \nR^3 \times \cX \to [0, 1]
  \label{eq:occupancy_network}
\end{equation}%
We call this network the \emph{Occupancy Network}.

\subsection{Training}\label{sec:method-training}

To learn the parameters $\theta$ of the neural network $f_\theta(p,x)$, we randomly sample points in the 3D bounding volume of the object under consideration:
for the $i$-th sample in a training batch we sample $K$ points $p_{ij}\in\nR^3$, $j=1,\dots,K$.
We then evaluate the mini-batch loss $\cL_\cB$ at those locations:
\begin{equation}\label{eq:objective}
\cL_{\cB}(\theta) 
= \frac{1}{|\cB|}\sum_{i=1}^{|\cB|} \sum_{j=1}^ K \cL(f_\theta(p_{ij},x_i), o_{ij})
\end{equation}
Here, $x_i$ is the $i$'th observation of batch $\cB$, $o_{ij} \equiv o(p_{ij})$ denotes the true occupancy at point $p_{ij}$, and $\cL(\cdot,\cdot)$ is a cross-entropy classification loss.

The performance of our method depends on the sampling scheme that we employ for drawing the locations $p_{ij}$ that are used for training.
In Section~\ref{sec:ablation-study} we perform a detailed ablation study comparing different sampling schemes.
In practice, we found that sampling uniformly inside the bounding box of the object with an additional small padding yields the best results.

Our 3D representation can also be used for learning probabilistic latent variable models.
Towards this goal, we introduce an encoder network $g_\psi(\cdot)$
that takes locations $p_{ij}$ and occupancies $o_{ij}$ as input and predicts mean $\mu_\psi$
and standard deviation $\sigma_\psi$
of a Gaussian distribution $q_\psi(z | (p_{ij}, o_{ij})_{j=1:K})$ on latent $z\in \nR^L$ as output.
We optimize a lower bound \cite{Kingma2014ICLR,Rezende2014ICML,Garnelo2018ARXIV} to the negative log-likelihood of the generative model $p((o_{ij})_{j=1:K} | (p_{ij})_{j=1:K})$:
\begin{multline}\label{eq:objective-gen}
\cL^{\text{gen}}_{\cB}(\theta, \psi) 
= \frac{1}{|\cB|}  \sum_{i=1}^{|\cB|} 
  \Bigl[\sum_{j=1}^ K \cL(f_\theta(p_{ij},z_i), o_{ij}) \\
+ \KL \left(q_\psi(z | (p_{ij}, o_{ij})_{j=1:K}) \,\|\, p_0(z) \right)
\Bigr]
\end{multline}
where $\KL$ denotes the KL-divergence, $p_0(z)$ is a prior distribution on the latent variable $z_i$ (typically Gaussian) and $z_i$ is sampled according to  $q_\psi(z_i | (p_{ij}, o_{ij})_{j=1:K})$.

\subsection{Inference}
\begin{figure}
\centering
\includegraphics[width=\linewidth]{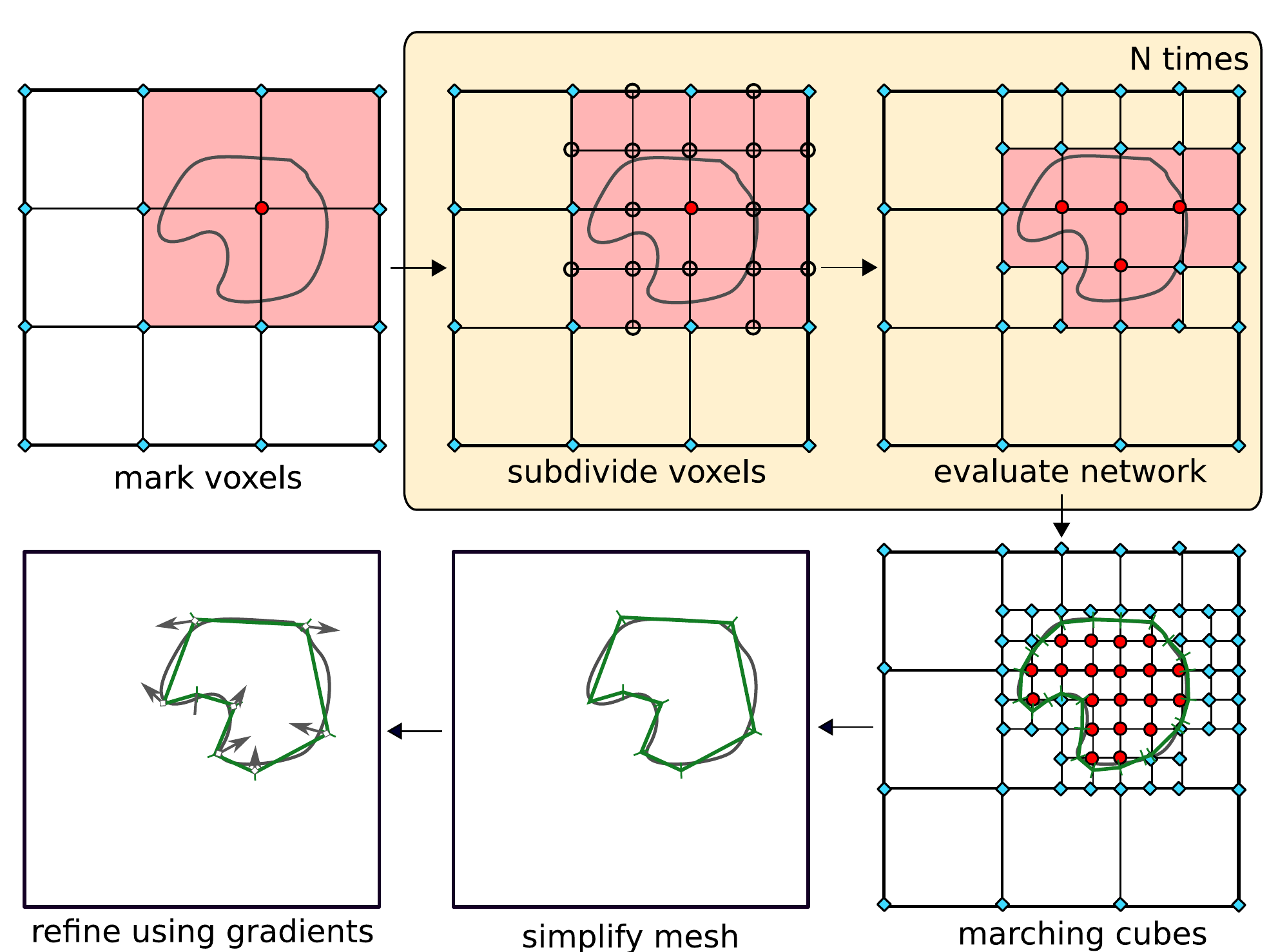}
\caption{%
\textbf{Multiresolution IsoSurface Extraction:}
We first mark all points at a given resolution which have already been evaluated as either occupied (red circles) or unoccupied (cyan diamonds).
We then determine all voxels that have both occupied and unoccupied corners and mark them as active (light red) and subdivide them into 8 subvoxels each.
Next, we evaluate all new grid points (empty circles) that have been introduced by the subdivision.
The previous two steps are repeated until the desired output resolution is reached.
Finally we extract the mesh using the marching cubes algorithm \cite{Lorensen1987SIGGRAPH}, simplify and refine the output mesh using first and second order gradient information.
}
\label{fig:mesh-extraction}
\vspace{-0.5cm}
\end{figure}
For extracting the isosurface corresponding to a new observation given a trained occupancy network, we introduce \emph{Multiresolution IsoSurface Extraction (MISE)},
a hierarchical isosurface extraction algorithm (\figref{fig:mesh-extraction}).
By incrementally building an octree~\cite{Jackins1980,Meagher1982,Szeliski1993CVGIP,Wong2001ECCV},
MISE enables us to extract high resolution meshes from the occupancy network  without densely evaluating all points of a high-dimensional occupancy grid.

We first discretize the volumetric space at an initial resolution and evaluate the occupancy network $f_\theta(p,x)$ for all $p$ in this grid.
We mark all grid points $p$ as occupied for which $f_\theta(p,x)$ is bigger or equal to some threshold\footnote{%
The threshold $\tau$ is the only hyperparameter of our occupancy network. It determines the ``thickness'' of the extracted 3D surface. In our experiments we cross-validate this threshold on a validation set.}
$\tau$.
Next, we mark all voxels as active for which at least two adjacent grid points have differing occupancy predictions.
These are the voxels which would intersect the mesh if we applied the marching cubes algorithm at the current resolution.
We subdivide all active voxels into 8 subvoxels and evaluate all new grid points which are introduced to the occupancy grid through this subdivision.
We repeat these steps until the desired final resolution is reached.
At this final resolution, we apply the Marching Cubes algorithm \cite{Lorensen1987SIGGRAPH} to extract an approximate isosurface
\begin{equation}
\{p \in \mathbb R^3 \mid f_\theta(p, x) = \tau \}.
\end{equation}
Our algorithm converges to the correct mesh if
the occupancy grid at the initial resolution contains points from every connected component of both the interior and the exterior of the mesh.
It is hence important to take an initial resolution which is high enough to satisfy this condition.
In practice, we found that an initial resolution of $32^3$ was sufficient in almost all cases.

The initial mesh extracted by the Marching Cubes algorithm can be further refined.
In a first step, we simplify the mesh using the Fast-Quadric-Mesh-Simplification algorithm\footnote{%
\url{https://github.com/sp4cerat/Fast-Quadric-Mesh-Simplification}
}~\cite{Garland1998VP}.
Finally, we refine the output mesh using first and second order (\ie, gradient) information.
Towards this goal, we sample random points $p_k$ from each face of the output mesh and minimize the loss
\begin{equation}\label{eq:refinement-loss}
 \sum_{k=1}^ K (f_\theta(p_k, x) - \tau)^2 + 
 \lambda \left\|
 \frac{\nabla_{p}f_\theta(p_k, x)}
  {{\|\nabla_{p}f_\theta(p_k, x)\|}}
  - n(p_k)
 \right\|^2
\end{equation}
where $n(p_k)$ denotes the normal vector of the mesh at $p_k$. In practice, we set $\lambda=0.01$.
Minimization of the second term in \eqref{eq:refinement-loss} uses second order gradient information and can be efficiently implemented using Double-Backpropagation~\cite{Drucker1992TNN}.

Note that this last step removes the discretization artifacts of the Marching Cubes approximation
and would not be possible if we had directly predicted a voxel-based representation.
In addition, our approach also allows to efficiently extract normals
for all vertices of our output mesh by simply backpropagating through
the occupancy network.
In total, our inference algorithm requires 3s per mesh.

\subsection{Implementation Details}\label{sec:method-implementation}

We implemented our occupancy network using  a fully-connected neural network with 5 ResNet blocks~\cite{He2016CVPR} and condition it on the input using conditional batch normalization \cite{Vries2017NIPS, Dumoulin2017ICLR}.
We exploit different encoder architectures depending on the type of input.
For single~view 3D~reconstruction, we use a 
ResNet18 architecture~\cite{He2016CVPR}.
For point clouds
we use the PointNet encoder~\cite{Qi2017CVPR}. 
For voxelized inputs, we use a 3D convolutional neural network~\cite{Maturana2015IROS}.
For unconditional mesh generation, we use a PointNet~\cite{Qi2017CVPR} for the encoder network $g_\psi$.
More details are provided in the supplementary material.

%% file: parts/experiments.tex
\section{Experiments}

We conduct three types of experiments to validate the proposed occupancy networks. 
First, we analyze the \textbf{representation power} of occupancy networks by examining how well the network can reconstruct complex 3D shapes from a learned latent embedding.
This gives us an upper bound on the results we can achieve when conditioning our representation on additional input.
Second, we \textbf{condition} our occupancy networks on images, noisy point clouds and low resolution voxel representations, and compare the performance of our method to several state-of-the-art baselines.
Finally, we examine the \textbf{generative} capabilities of occupancy networks by adding an encoder to our model and generating unconditional samples from this model.\footnote{%
The code to reproduce our experiments is available under \url{https://github.com/LMescheder/Occupancy-Networks}.
}

\boldparagraph{Baselines}
For the single image 3D reconstruction task, we compare our approach against several state-of-the-art baselines which leverage various 3D representations:
we evaluate against 3D-R2N2 \cite{Choy2016ECCV} as a voxel-based method, Point Set Generating Networks
(PSGN) \cite{Fan2017CVPR} as a point-based technique and Pixel2Mesh \cite{Wang2018ECCV} as well as AtlasNet~\cite{Groueix2018CVPR} as mesh-based approaches.
For point cloud inputs, we adapted 3D-R2N2 and PSGN by changing the encoder.
As mesh-based baseline, we use Deep Marching Cubes (DMC) \cite{Liao2018CVPR} which has recently reported state-of-the-art results on this task.
For the voxel super-resolution task we assess the improvements \wrt the input.

\boldparagraph{Dataset}
For all of our experiments we use the ShapeNet \cite{Chang2015ARXIV} subset of Choy \etal \cite{Choy2016ECCV}.
We also use the same voxelization, image renderings and train/test split as Choy \etal.
Moreover, we subdivide the training set into a training and a validation set on which we track the loss of our method and the baselines to determine when to stop training.

In order to generate watertight meshes and to determine if a point lies in the interior of a mesh (\eg, for measuring IoU)
we use the code provided by Stutz \etal \cite{Stutz2018CVPR}.
For a fair comparison, we sample points from the surface of the watertight mesh instead of the original model as ground truth for PSGN~\cite{Fan2017CVPR}, Pixel2Mesh~\cite{Wang2018ECCV} and DMC~\cite{Liao2018CVPR}.
All of our evaluations are conducted \wrt these watertight meshes.

\boldparagraph{Metrics}
For evaluation we use the volumetric IoU, the Chamfer-$L_1$ distance and a normal consistency score. 

Volumetric IoU is defined as the quotient of the volume of the two meshes' union and the volume of their intersection.
We obtain unbiased estimates of the volume of the intersection and the union by randomly sampling 100k points from the bounding volume and determining if the points lie inside our outside the ground truth / predicted mesh.

The Chamfer-$L_1$ distance is defined as the mean of an accuracy and and a completeness metric.
The accuracy metric is defined as the mean distance of points on the output mesh to their nearest neighbors on the ground truth mesh.
The completeness metric is defined similarly, but in opposite direction.
We estimate both distances efficiently by randomly sampling 100k points from both meshes and using a KD-tree to estimate the corresponding distances.
Like Fan~\etal~\cite{Fan2017CVPR} we use $1/10$ times the maximal edge length of the current object's bounding box as unit $1$.

Finally, to measure how well the methods can capture higher order information, we define a normal consistency score as the mean absolute dot product of the normals in one mesh and the normals at the corresponding nearest neighbors in the other mesh.

\subsection{Representation Power}\label{sec:experiments-repr-power}
\begin{figure}
 \centering
 \begin{tabular}{@{}c@{}c@{}c@{}c|c}
\includegraphics[width=0.19\linewidth]{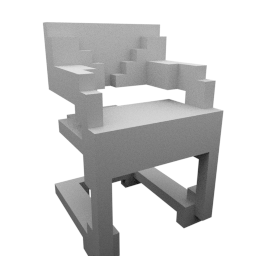} &
\includegraphics[width=0.19\linewidth]{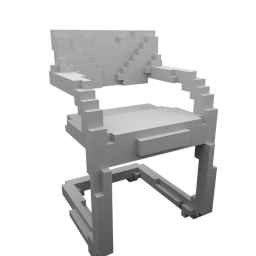} &
\includegraphics[width=0.19\linewidth]{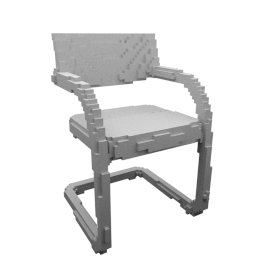} & 
\includegraphics[width=0.19\linewidth]{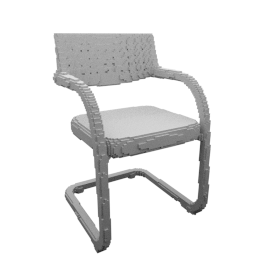} &
\includegraphics[width=0.19\linewidth]{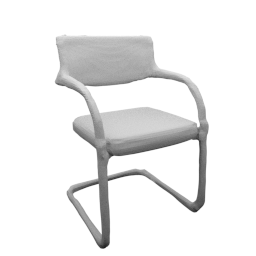} \\
\;$16^3$ & \;$32^3$ & \;$64^3$ & \;$128^3$ & \;ours
\end{tabular}
\caption{
\textbf{Discrete vs. Continuous.}
Qualitative comparison of our continuous representation (right) to voxelizations at various resolutions (left).
Note how our representation encodes details which are lost in 
voxel-based representations.}
\label{fig:qualitative-voxels}
\vspace{-0.5cm}
\end{figure}

\begin{figure}
\centering
\includegraphics[width=\linewidth]{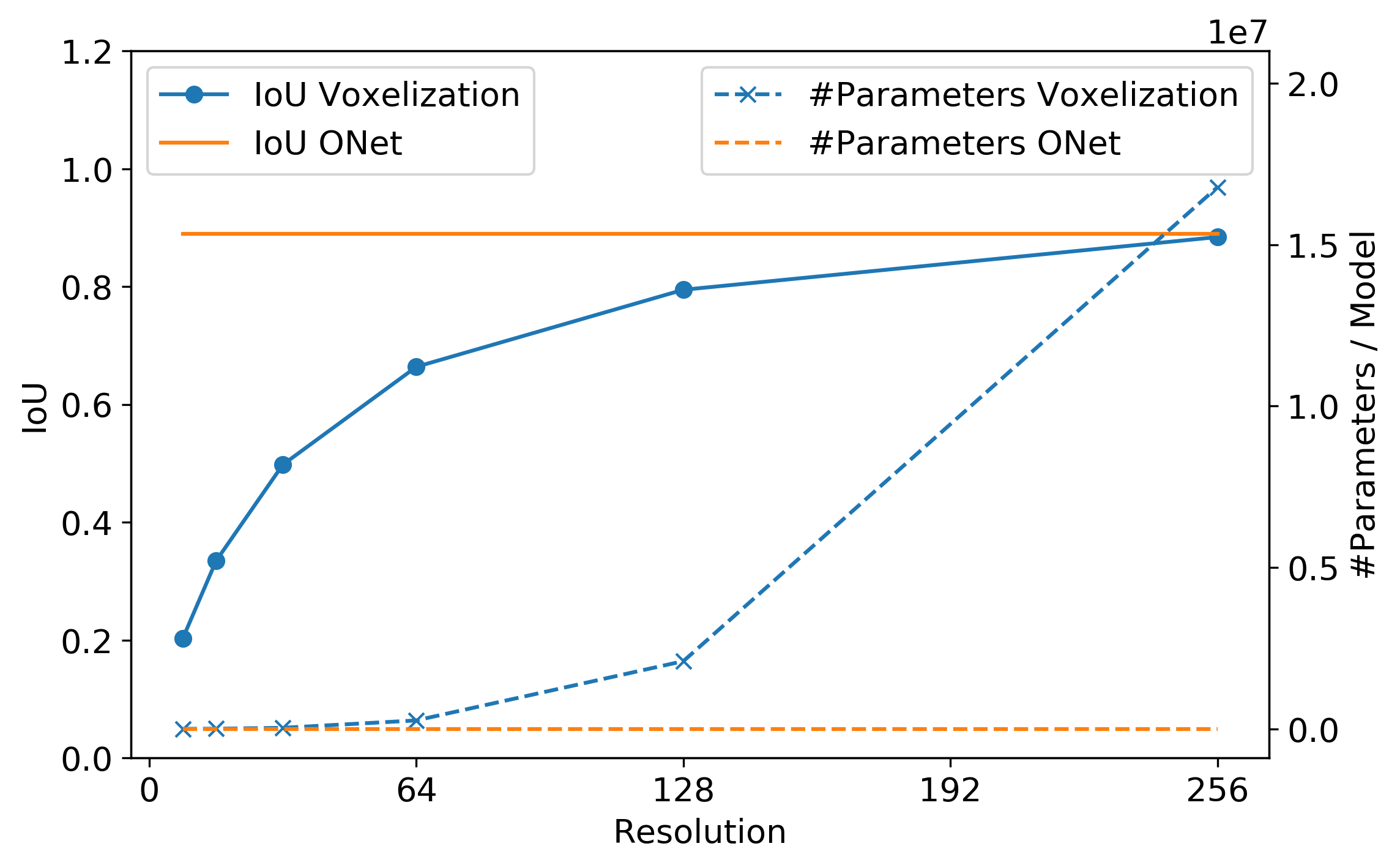}
\caption{
\textbf{IoU vs. Resolution.} 
This plot shows the IoU of a voxelization to the ground truth mesh (solid blue line) in comparison to our continuous representation (solid orange line)
as well as the number of parameters per model needed for the two representations (dashed lines).
Note how our representation leads to larger IoU \wrt the ground truth mesh compared to a low-resolution voxel representation.
At the same time, the number of parameters of a voxel representation grows cubically with the resolution, whereas the number of parameters of occupancy networks is independent of the resolution.
}
\label{fig:idx-to-mesh}
\vspace{-0.5cm}
\end{figure}

In our first experiment, we investigate how well occupancy networks represent 3D geometry, independent of the inaccuracies of the input encoding.
The question we try to answer in this experiment is whether our network can learn a memory efficient representation of 3D shapes while at the same time preserving as many details as possible.
This gives us an estimate of the representational capacity of our model and an upper bound on the performance we may expect when conditioning our model on additional input.
Similarly to \cite{Tatarchenko2017ICCV},
we embed each training sample in a 512 dimensional latent space and train our neural network to 
reconstruct the 3D shape from this embedding.

We apply our method to the training split of the ``chair'' category of the ShapeNet dataset.
This subset is challenging to represent as it is highly varied and many models contain high-frequency details.
Since we are only interested in reconstructing the training data, we do not use separate validation and test sets for this experiment.

For evaluation, we measure the volumetric IoU to the ground truth mesh. 
Quantitative results and a comparison to voxel representations at various resolutions are shown in \figref{fig:idx-to-mesh}.
We see that the Occupancy Network (ONet) is able to faithfully represent the entire dataset with a high mean IoU of 0.89 while a
low-resolution voxel representation is not able to represent the meshes accurately.
At the same time, the occupancy network is able to encode all 4746 training samples with as little as ~6M parameters, independently of the resolution.
In contrast, the memory requirements of a voxel representation grow cubically with resolution.
Qualitative results are shown in \figref{fig:qualitative-voxels}.
We observe that the occupancy network enables us to represent details of the 3D geometry which are lost in a low-resolution voxelization.

\subsection{Single Image 3D Reconstruction}\label{sec:experiments-im2mesh}

In our second experiment, 
we condition the occupancy network on an additional view of the object from a random camera location.
The goal of this experiment is to evaluate how well occupancy functions can be inferred from complex input.
While we train and test our method on the ShapeNet dataset, we also present qualitative results for
the KITTI \cite{Geiger2013IJRR} and the Online Products dataset \cite{Oh2016CVPR}.

\begin{table*}
\centering
\resizebox{\textwidth}{!}{%
\input{tables/table_img2mesh.tex}
}\caption{
\textbf{Single Image 3D Reconstruction.}
This table shows a numerical comparison of our approach and the baselines for single image 3D reconstruction on the ShapeNet dataset.
We measure the IoU, Chamfer-$L_1$ distance and Normal Consistency for various methods \wrt the ground truth mesh.
Note that in contrast to prior work, we compute the IoU \wrt the high-resolution mesh and not a coarse voxel representation. 
All methods apart from AtlasNet \cite{Groueix2018CVPR} are evaluated on the test split by Choy \etal \cite{Choy2016ECCV}. Since AtlasNet uses a pretrained model, we evaluate it on the intersection of the test splits from \cite{Choy2016ECCV} and \cite{Groueix2018CVPR}.
}
\label{tab:img-to-mesh-all-classes}
\vspace{-0.3cm}
\end{table*}

\begin{figure}
\centering
\setlength{\tabcolsep}{0pt}
\def\arraystretch{0}
\begin{tabular}{@{}c@{}c@{}c@{}c@{}c@{}c}
\footnotesize{Input} & 
\footnotesize{3D-R2N2}  & 
\footnotesize{PSGN} & 
\footnotesize{Pix2Mesh} &
\footnotesize{AtlasNet}  &
\footnotesize{Ours} \\
\includegraphics[width=0.16\linewidth]{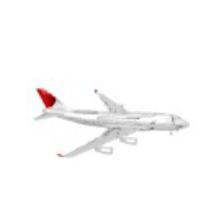} &
\includegraphics[width=0.16\linewidth,trim={1.5cm 1.8cm 1.5cm 1.8cm},clip]{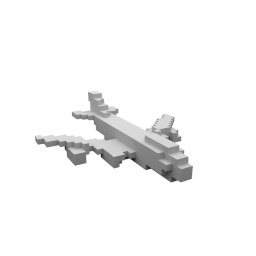} &
\includegraphics[width=0.16\linewidth,trim={1.5cm 1.8cm 1.5cm 1.8cm},clip]{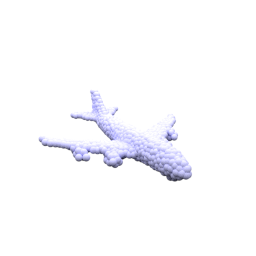} &
\includegraphics[width=0.16\linewidth,trim={1.5cm 1.8cm 1.5cm 1.8cm},clip]{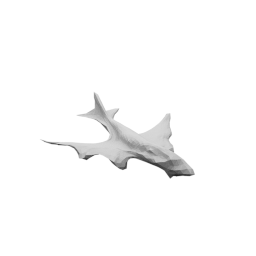} &
\includegraphics[width=0.16\linewidth,trim={1.5cm 1.8cm 1.5cm 1.8cm},clip]{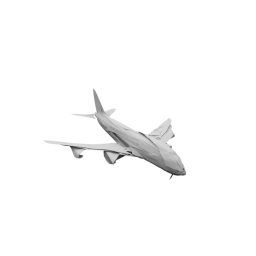} &
\includegraphics[width=0.16\linewidth,trim={1.5cm 1.8cm 1.5cm 1.8cm},clip]{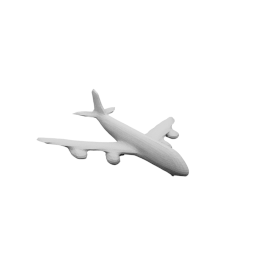} \\
\includegraphics[width=0.16\linewidth]{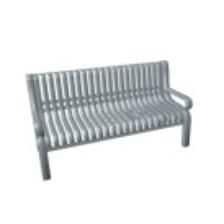} &
\includegraphics[width=0.16\linewidth,trim={.8cm .8cm .8cm .8cm},clip]{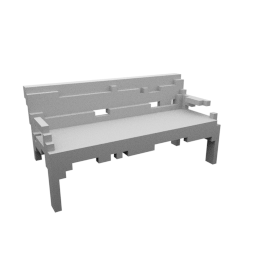} &
\includegraphics[width=0.16\linewidth,trim={.8cm .8cm .8cm .8cm},clip]{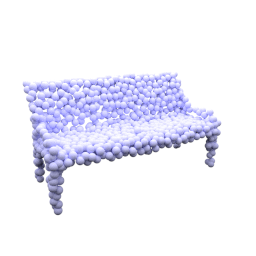} &
\includegraphics[width=0.16\linewidth,trim={.8cm .8cm .8cm .8cm},clip]{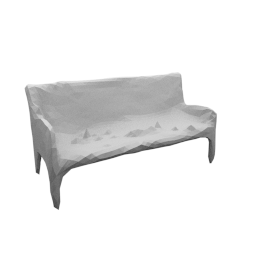} &
\includegraphics[width=0.16\linewidth,trim={.8cm .8cm .8cm .8cm},clip]{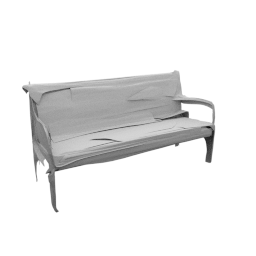} &
\includegraphics[width=0.16\linewidth,trim={.8cm .8cm .8cm .8cm},clip]{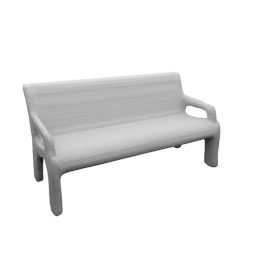} \\
\includegraphics[width=0.16\linewidth]{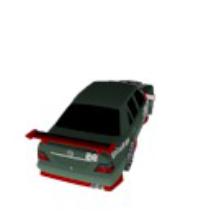} &
\includegraphics[width=0.16\linewidth,trim={1.5cm 1.5cm 1.5cm 1.5cm},clip]{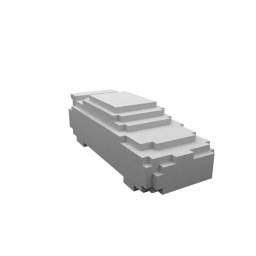} &
\includegraphics[width=0.16\linewidth,trim={1.5cm 1.5cm 1.5cm 1.5cm},clip]{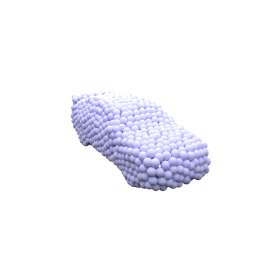} &
\includegraphics[width=0.16\linewidth,trim={1.5cm 1.5cm 1.5cm 1.5cm},clip]{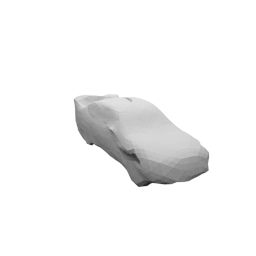} &
\includegraphics[width=0.16\linewidth,trim={1.5cm 1.5cm 1.5cm 1.5cm},clip]{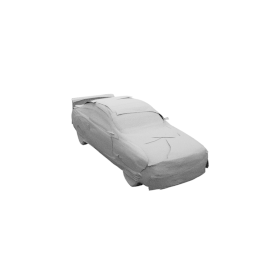} &
\includegraphics[width=0.16\linewidth,trim={1.5cm 1.5cm 1.5cm 1.5cm},clip]{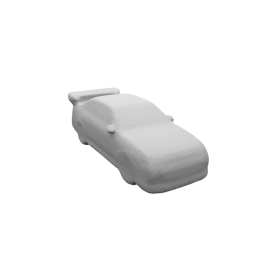} \\
\includegraphics[width=0.16\linewidth]{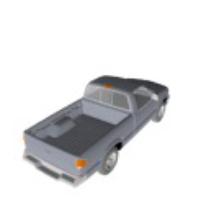} &
\includegraphics[width=0.16\linewidth,trim={1.5cm 1.5cm 1.5cm 1.5cm},clip]{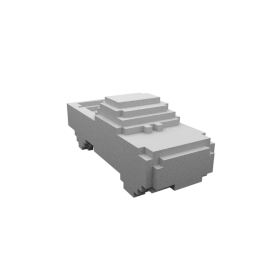} &
\includegraphics[width=0.16\linewidth,trim={1.5cm 1.5cm 1.5cm 1.5cm},clip]{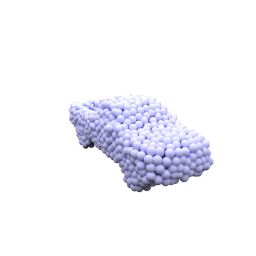} &
\includegraphics[width=0.16\linewidth,trim={1.5cm 1.5cm 1.5cm 1.5cm},clip]{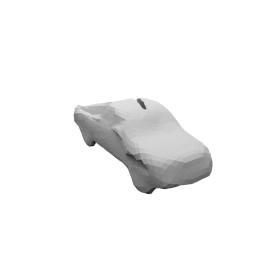} &
\includegraphics[width=0.16\linewidth,trim={1.5cm 1.5cm 1.5cm 1.5cm},clip]{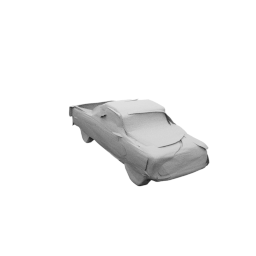} &
\includegraphics[width=0.16\linewidth,trim={1.5cm 1.5cm 1.5cm 1.5cm},clip]{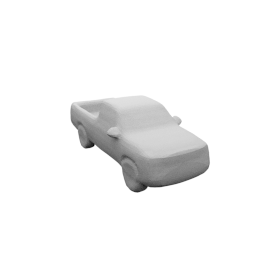} \\
\includegraphics[width=0.16\linewidth]{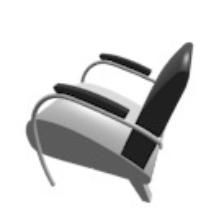} &
\includegraphics[width=0.16\linewidth]{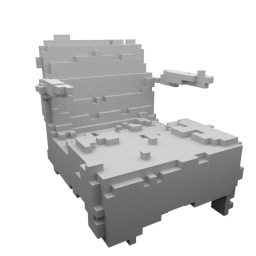} &
\includegraphics[width=0.16\linewidth]{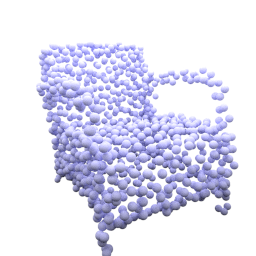} &
\includegraphics[width=0.16\linewidth]{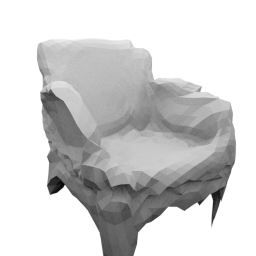} &
\includegraphics[width=0.16\linewidth]{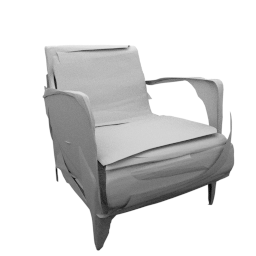} &
\includegraphics[width=0.16\linewidth]{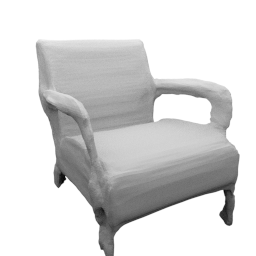} \\
\includegraphics[width=0.16\linewidth]{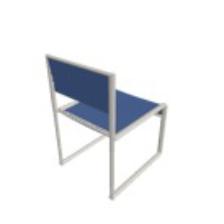} &
\includegraphics[width=0.16\linewidth]{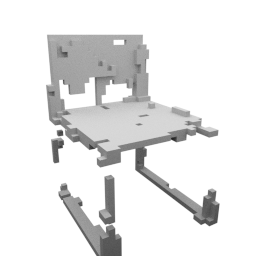} &
\includegraphics[width=0.16\linewidth]{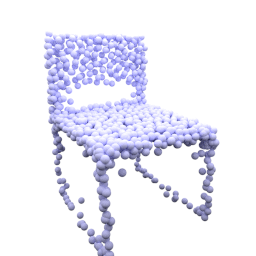} &
\includegraphics[width=0.16\linewidth]{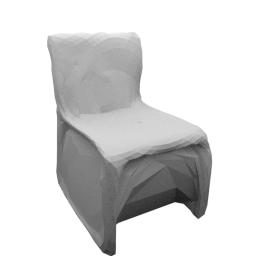} &
\includegraphics[width=0.16\linewidth]{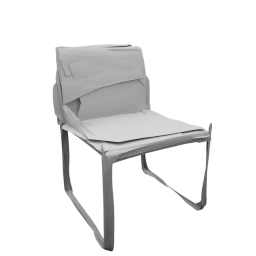} &
\includegraphics[width=0.16\linewidth]{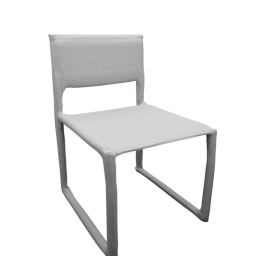} \\
\includegraphics[width=0.16\linewidth]{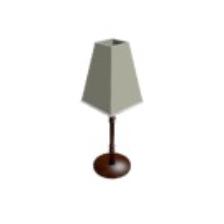} &
\includegraphics[width=0.16\linewidth,trim={.5cm .5cm .5cm .5cm},clip]{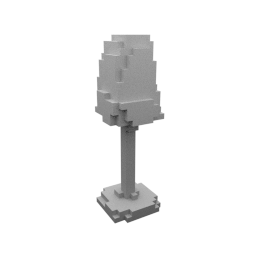} &
\includegraphics[width=0.16\linewidth,trim={.5cm .5cm .5cm .5cm},clip]{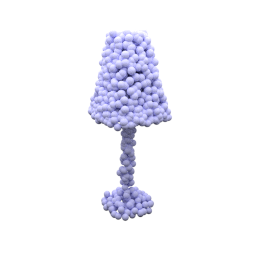} &
\includegraphics[width=0.16\linewidth,trim={.5cm .5cm .5cm .5cm},clip]{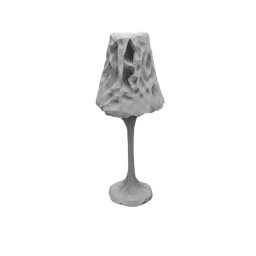} &
\includegraphics[width=0.16\linewidth,trim={.5cm .5cm .5cm .5cm},clip]{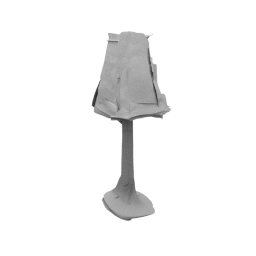} &
\includegraphics[width=0.16\linewidth,trim={.5cm .5cm .5cm .5cm},clip]{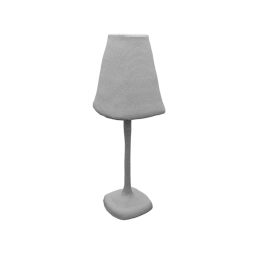} \\
\includegraphics[width=0.16\linewidth]{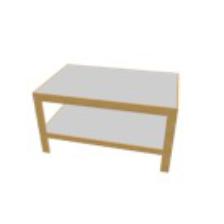} &
\includegraphics[width=0.16\linewidth,trim={.5cm .5cm .5cm .5cm},clip]{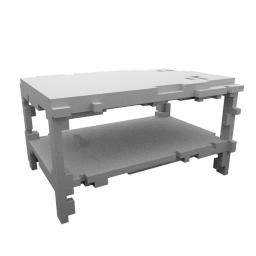} &
\includegraphics[width=0.16\linewidth,trim={.5cm .5cm .5cm .5cm},clip]{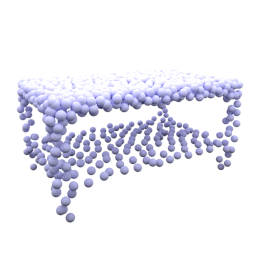} &
\includegraphics[width=0.16\linewidth,trim={.5cm .5cm .5cm .5cm},clip]{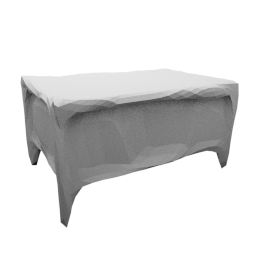} &
\includegraphics[width=0.16\linewidth,trim={.5cm .5cm .5cm .5cm},clip]{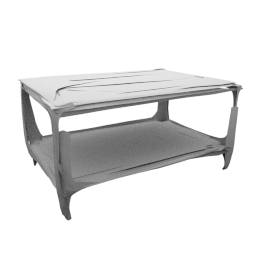} &
\includegraphics[width=0.16\linewidth,trim={.5cm .5cm .5cm .5cm},clip]{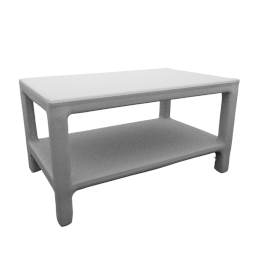}
\end{tabular}
\caption{
\textbf{Single Image 3D Reconstruction.}
The input image is shown in the first column, the other columns show 
the results for our method compared to various baselines.
}
\label{fig:qualitative-results}
\vspace{-0.5cm}
\end{figure}

\boldparagraph{ShapeNet}
In this experiment, we use a ResNet-18 image encoder, which was pretrained on the ImageNet dataset.
For a fair comparison, we use the same image encoder for both 3D-R2N2 and PSGN\footnote{%
See supplementary for a comparison to the original architectures.}.
For PSGN we use a fully connected decoder with 4 layers and 512 hidden units in each layer.
The last layer projects the hidden representation to a 3072 dimensional vector which we reshape into 1024 3D points.
As we use only a single input view, we remove the recurrent network in 3D-R2N2.
We reimplemented the method of \cite{Wang2018ECCV} in PyTorch, closely following the Tensorflow implementation provided by the authors.
For the method of \cite{Groueix2018CVPR}, we use the code and pretrained model from the authors\footnote{\url{https://github.com/ThibaultGROUEIX/AtlasNet}}.

For all methods, we track the loss and other metrics on the validation set and stop training as soon as the target metric reaches its optimum.
For 3D-R2N2 and our method we use the IoU to the ground truth mesh as target metric, for PSGN and Pixel2Mesh we use the Chamfer distance to the
ground truth mesh as target metric. To extract the final mesh, we use a threshold of $0.4$ for 3D-R2N2 as suggested in the original publication \cite{Choy2016ECCV}.
To choose the threshold parameter $\tau$ for our method, we perform grid search on the validation set
(see supplementary) and found that $\tau = 0.2$ yields a good trade-off between accuracy and completeness.

Qualitative results from our model and the baselines are shown in \figref{fig:qualitative-results}. 
We observe that all methods are able to capture the 3D geometry of the input image.
However, 3D-R2N2 produces a very coarse representation and hence lacks details.
In contrast, PSGN produces a high-fidelity output, but lacks connectivity.
As a result, PSGN requires additional lossy post-processing steps to produce a final mesh\footnote{See supplementary material for meshing results.}.
Pixel2Mesh is able to create compelling meshes, but often misses holes in the presence of more complicated topologies.
Such topologies are frequent, for example, for the ``chairs`` category in the ShapeNet dataset.
Similarly, AtlasNet captures the geometry well, but produces artifacts in form of self-intersections and overlapping patches.

In contrast, our method is able to capture complex topologies, produces closed meshes and preserves most of the details.
Please see the supplementary material for additional high resolution results and failure cases.

Quantitative results are shown in Table~\ref{tab:img-to-mesh-all-classes}. 
We observe that our method achieves the highest IoU and normal consistency to the ground truth mesh.
Surprisingly, while not trained \wrt Chamfer distance as PSGN, Pixel2Mesh or AtlasNet, our method also achieves good results
for this metric.
Note that it is not possible to evaluate the IoU for PSGN or AtlasNet, as they do not yield watertight meshes.

\begin{figure}
\centering
\setlength{\tabcolsep}{0pt}
\def\arraystretch{0}
\begin{subfigure}{0.45\linewidth}
\centering
\begin{tabular}{@{}c@{}c@{}c}
\small{Input} & \quad & \small{Reconstruction} \\
\includegraphics[width=0.35\linewidth]{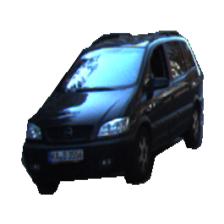} &&
\includegraphics[width=0.35\linewidth,trim={1cm 1cm 1cm 1cm},clip]{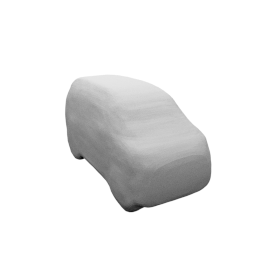} \\
\includegraphics[width=0.35\linewidth]{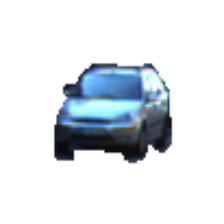} &&
\includegraphics[width=0.35\linewidth,trim={1cm 1cm 1cm 1cm},clip]{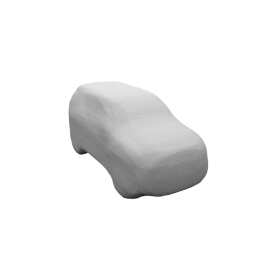} \\
\includegraphics[width=0.35\linewidth]{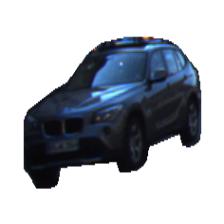} &&
\includegraphics[width=0.35\linewidth,trim={1cm 1cm 1cm 1cm},clip]{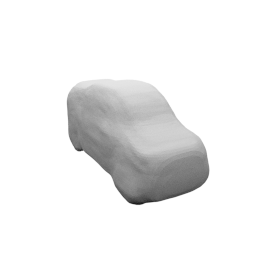} \\
\includegraphics[width=0.35\linewidth]{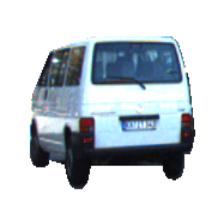} &&
\includegraphics[width=0.35\linewidth,trim={1cm 1cm 1cm 1cm},clip]{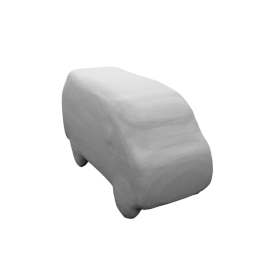} \\
\end{tabular}
\caption{KITTI}
\label{fig:real-data-kitti}
\end{subfigure}
\quad
\begin{subfigure}{0.45\linewidth}
\centering
\begin{tabular}{@{}c@{}c@{}c}
\small{Input} & \quad & \small{Reconstruction} \\
\includegraphics[width=0.35\linewidth]{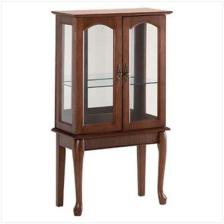} &&
\includegraphics[width=0.35\linewidth]{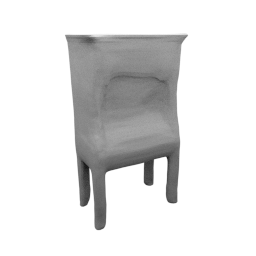} \\
\includegraphics[width=0.35\linewidth]{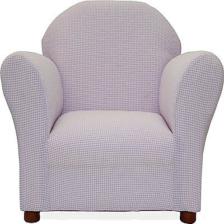} &&
\includegraphics[width=0.35\linewidth]{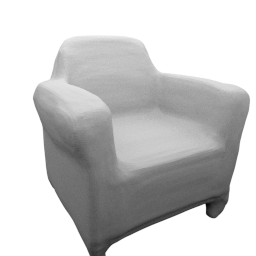} \\
\includegraphics[width=0.35\linewidth]{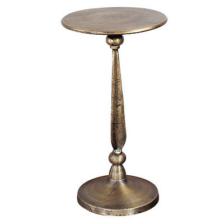} &&
\includegraphics[width=0.35\linewidth]{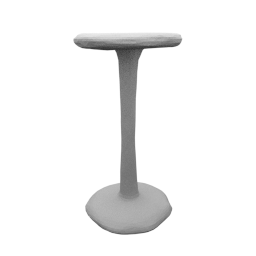} \\
\includegraphics[width=0.35\linewidth]{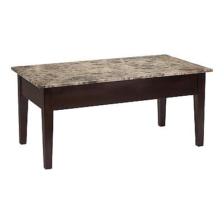} &&
\includegraphics[width=0.35\linewidth]{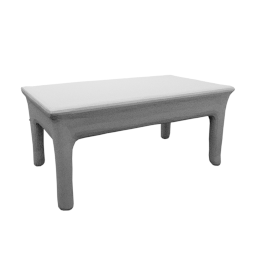} \\
\end{tabular}
\caption{Online Products}
\label{fig:real-data-online-products}
\end{subfigure}
\caption{\textbf{Qualitative results for real data.}
We applied our trained model to the KITTI and Online Products datasets.
Despite only trained on synthetic data, our model generalizes reasonably well to real data.
}
\label{fig:real-data-oproducts}
\label{fig:real-data}
\vspace{-0.5cm}
\end{figure}

\boldparagraph{Real Data}
To test how well our model generalizes to real data, we  
apply our network to 
the KITTI \cite{Geiger2013IJRR}  and Online Products datasets \cite{Oh2016CVPR}.
To capture the variety in viewpoints of KITTI and Online Products,
we rerendered all ShapeNet objects with random camera locations
and retrained our network for this task.

For the KITTI dataset, we additionally use the instance masks provided in \cite{Alhaija2017BMVC}
to mask and crop car regions.
We then feed these images into our neural network to predict the occupancy function.
Some selected qualitative results are shown in \figref{fig:real-data-kitti}.
Despite only trained on synthetic data, we observe that our method is also able to 
generate realistic reconstructions in this challenging setting.

For the Online Products dataset, we apply the same pretrained model.
Several qualitative results are shown in \figref{fig:real-data-online-products}.
Again, we observe that our method generalizes reasonably well to real images
despite being trained solely on synthetic data.
An additional quantitative evaluation on the Pix3D dataset~\cite{Sun2018CVPR_pix3d} can be found in the supplementary.

\subsection{Point Cloud Completion}

As a second conditional task, we apply our method to the problem of reconstructing the mesh from noisy point clouds.
Towards this goal, we subsample 300 points from the surface of each of the (watertight) ShapeNet models and apply noise
using a Gaussian distribution with zero mean and standard deviation $0.05$ to the point clouds.

Again, we measure both the IoU and Chamfer-$L_1$ distance \wrt the ground truth mesh.
The results are shown in Table~\ref{tab:pointcloud}.
We observe that our method achieves the highest IoU and normal consistency as well as the lowest Chamfer-$L_1$ distance.
Note that all numbers are significantly better than for the single image 3D reconstruction task.
This can be explained by the fact that this task is much easier for the recognition model, as there is less ambiguity and the model only has to fill in the gaps.

\subsection{Voxel Super-Resolution}

As a final conditional task, we apply occupancy networks to 3D super-resolution~\cite{Smith2018NIPS}.
Here, the task is to reconstruct a high-resolution mesh from a coarse $32^3$ voxelization of this mesh.

The results are shown in Table~\ref{tab:voxel-superresolution}.
We observe that our model considerably improves IoU, Chamfer-$L_1$ distance and normal consistency  compared to the coarse input mesh.
Please see the supplementary for qualitative results.

\renewcommand*{\thefootnote}{\fnsymbol{footnote}}
\begin{table}
\centering
\resizebox{.9\linewidth}{!}{%
\input{tables/table_pcl2mesh_small.tex}
}
\caption{
\textbf{3D Reconstruction from Point Clouds.}
This table shows a numerical comparison of our approach \wrt the baselines for 3D reconstruction from point clouds on the ShapeNet dataset.
We measure IoU, Chamfer-$L_1$ distance and Normal Consistency \wrt the ground truth mesh.
}
\label{tab:pointcloud}
\vspace{-0.5cm}
\end{table}
\footnotetext[2]{
    Result for PSGN was corrected after CVPR camera-ready version.
}

\begin{table}
\centering
\resizebox{.85\linewidth}{!}{%
\input{tables/table_voxels2mesh_small.tex}
}
\caption{
\textbf{Voxel Super-Resolution.}
This table shows a numerical comparison of the output of our approach 
in comparison to the input on the ShapeNet dataset.
}
\label{tab:voxel-superresolution}
\vspace{-0.3cm}
\end{table}

\subsection{Unconditional Mesh Generation}

\begin{figure}
\centering
\setlength{\tabcolsep}{0pt}
\def\arraystretch{0}
\begin{tabular}{@{}c@{}c@{}c@{}c@{}c@{}c@{}c}
\includegraphics[width=0.14\linewidth,trim={.7cm .7cm .7cm .7cm},clip]{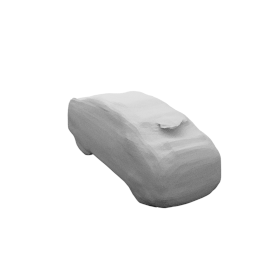} &
\includegraphics[width=0.14\linewidth,trim={.7cm .7cm .7cm .7cm},clip]{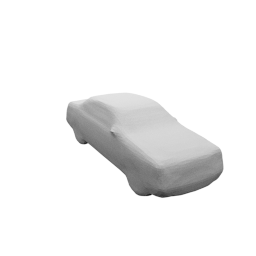} &
\includegraphics[width=0.14\linewidth,trim={.7cm .7cm .7cm .7cm},clip]{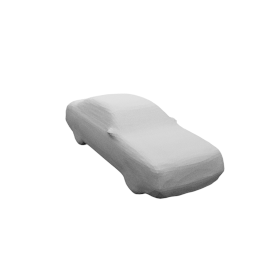} &
\includegraphics[width=0.14\linewidth,trim={.7cm .7cm .7cm .7cm},clip]{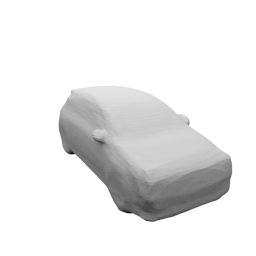} &
\includegraphics[width=0.14\linewidth,trim={.7cm .7cm .7cm .7cm},clip]{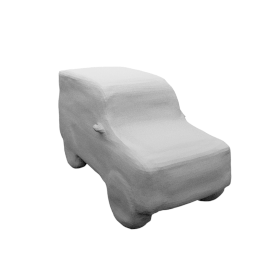} &
\includegraphics[width=0.14\linewidth,trim={.7cm .7cm .7cm .7cm},clip]{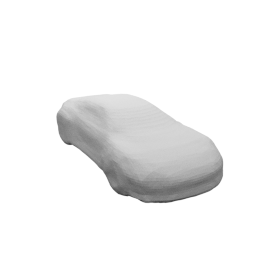} &
\includegraphics[width=0.14\linewidth,trim={.7cm .7cm .7cm .7cm},clip]{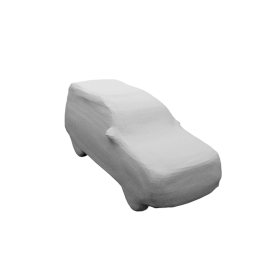} \\
\includegraphics[width=0.14\linewidth,trim={1.5cm 1.5cm 1.5cm 1.5cm},clip]{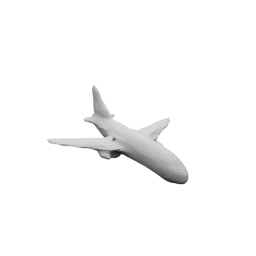} &
\includegraphics[width=0.14\linewidth,trim={1.5cm 1.5cm 1.5cm 1.5cm},clip]{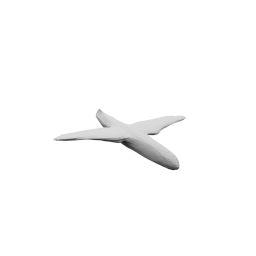} &
\includegraphics[width=0.14\linewidth,trim={1.5cm 1.5cm 1.5cm 1.5cm},clip]{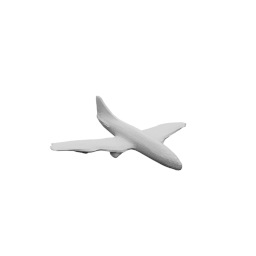} &
\includegraphics[width=0.14\linewidth,trim={1.5cm 1.5cm 1.5cm 1.5cm},clip]{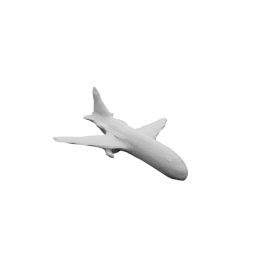} &
\includegraphics[width=0.14\linewidth,trim={1.5cm 1.5cm 1.5cm 1.5cm},clip]{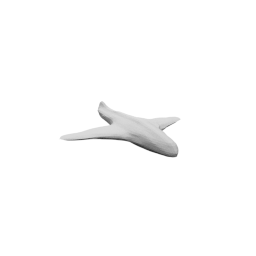} &
\includegraphics[width=0.14\linewidth,trim={1.5cm 1.5cm 1.5cm 1.5cm},clip]{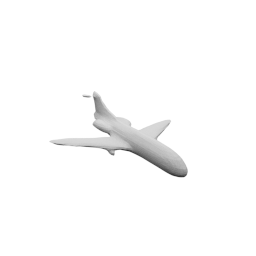} &
\includegraphics[width=0.14\linewidth,trim={1.5cm 1.5cm 1.5cm 1.5cm},clip]{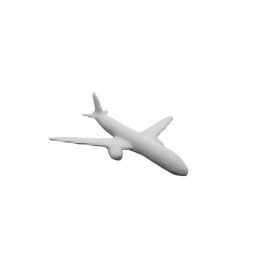} \\
\includegraphics[width=0.14\linewidth,trim={1cm 1cm 1cm 1cm},clip]{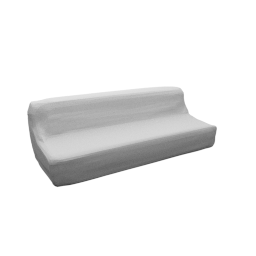} &
\includegraphics[width=0.14\linewidth,trim={1cm 1cm 1cm 1cm},clip]{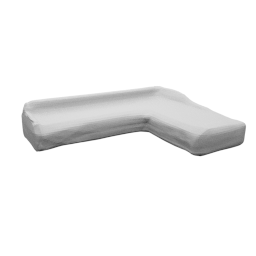} &
\includegraphics[width=0.14\linewidth,trim={1cm 1cm 1cm 1cm},clip]{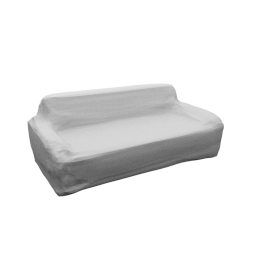} &
\includegraphics[width=0.14\linewidth,trim={1cm 1cm 1cm 1cm},clip]{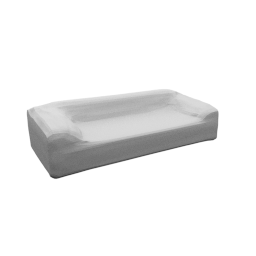} &
\includegraphics[width=0.14\linewidth,trim={1cm 1cm 1cm 1cm},clip]{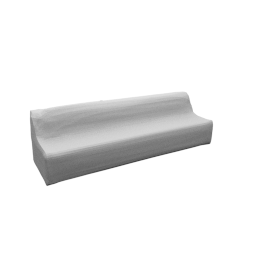} &
\includegraphics[width=0.14\linewidth,trim={1cm 1cm 1cm 1cm},clip]{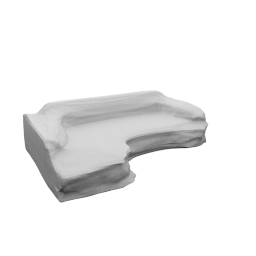} &
\includegraphics[width=0.14\linewidth,trim={1cm 1cm 1cm 1cm},clip]{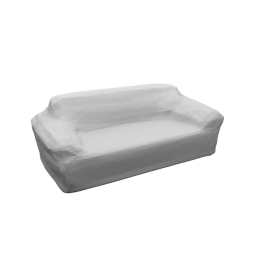} \\
\includegraphics[width=0.14\linewidth]{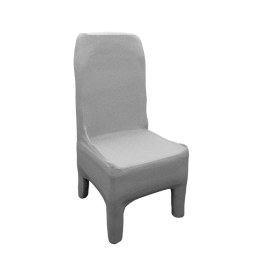} &
\includegraphics[width=0.14\linewidth]{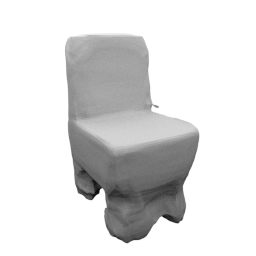} &
\includegraphics[width=0.14\linewidth]{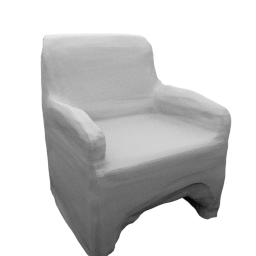} &
\includegraphics[width=0.14\linewidth]{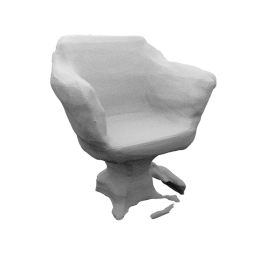} &
\includegraphics[width=0.14\linewidth]{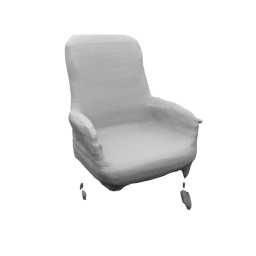} &
\includegraphics[width=0.14\linewidth]{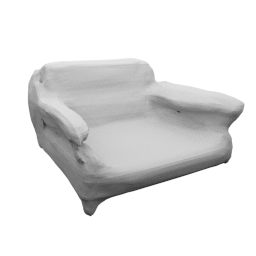} &
\includegraphics[width=0.14\linewidth]{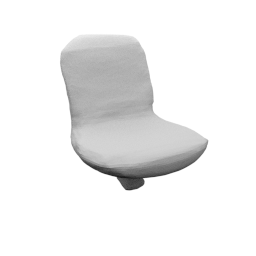} 
\end{tabular}

\caption{
\textbf{Unconditional 3D Samples.} Random samples of our unsupervised models trained on the categories ``car``, ``airplane``, ``sofa`` and ``chair`` of the ShapeNet dataset.
We see that our models are able to capture the distribution of 3D objects and produce compelling new samples.
}
\label{fig:cars-unconditional}
\vspace{-0.5cm}
\end{figure}

Finally, we apply our occupancy network to unconditional mesh generation, training it separately on four categories of the ShapeNet dataset in an unsupervised fashion.
Our goal is to explore how well our model can represent the latent space of 3D models.
Some samples are shown in Figure~\ref{fig:cars-unconditional}.
Indeed, we find that our model can generate compelling new models.
In the supplementary material we show interpolations in latent space for our model.

\subsection{Ablation Study}\label{sec:ablation-study}

In this section, we test how the various components of our model affect its performance
on the single-image 3D-reconstruction task.

\paragraph{Effect of sampling strategy}
First, we examine how the sampling strategy affects the performance of our final model.
We try three different sampling strategies:
(i) sampling 2048 points uniformly in the bounding volume of the ground truth mesh (uniform sampling),
(ii) sampling 1024 points inside and 1024 points outside mesh (equal sampling)
and
(iii) sampling 1024 points uniformly and 1024 points on the surface of the mesh plus some Gaussian noise with standard deviation $0.1$ (surface sampling).
We also examine the effect of the number of sampling points by decreasing this number from 2048 to 64.

The results are shown in Table~\ref{tab:ablation-sampling}.
To our surprise, we find that uniform, the simplest sampling strategy, works best.
We explain this by the fact that other sampling strategies introduce bias to the model:
for example, when sampling an equal number of points inside and outside the mesh, we implicitly tell the model that every object has a volume of $0.5$.
Indeed, when using this sampling strategy, we observe thickening artifacts in the model's output.
Moreover, we find that reducing the number of sampling points from 2048 to 64 still leads to good performance, although the model does not perform as well as a model trained with 2048 sampling points.

\paragraph{Effect of architecture}
To test the effect of the various components of our architecture, we test two variations:
(i) we remove the conditional batch normalization and replace it with a linear layer in the beginning of the network that projects the encoding of the input to the required hidden dimension and
(ii) we remove all ResNet blocks in the decoder and replace them with linear blocks.
The results are presented in Table~\ref{tab:ablation-architecture}.
We find that both components are helpful to achieve good performance.

\begin{table}
\centering
\begin{subfigure}{\linewidth}
\centering
\resizebox{.9\linewidth}{!}{%
\input{tables/ablations_sampling.tex}
}
\caption{Influence of Sampling Strategy}
\label{tab:ablation-sampling}
\end{subfigure}

\vspace{2ex}

\begin{subfigure}{\linewidth}
\centering
\resizebox{.9\linewidth}{!}{%
\input{tables/ablations_architecture.tex}
}
\caption{Influence of Occupancy Network Architecture}
\label{tab:ablation-architecture}
\end{subfigure}
\caption{%
\textbf{Ablation Study.}
When we vary the sampling strategy, we observe that uniform sampling in the bounding volume performs best.
Similarly, when we vary the architecture, we find that our ResNet architecture with conditional batch
normalization yields the best results.
}
\label{tab:ablation}
\vspace{-0.5cm}
\end{table}

%% file: tables/table_img2mesh.tex
\begin{tabular}{l|ccccc|ccccc|ccccc}
\toprule
{} & \multicolumn{5}{c}{IoU} & \multicolumn{5}{c}{Chamfer-$L_1$} & \multicolumn{5}{c}{Normal Consistency} \\
{} & 3D-R2N2 & PSGN &        Pix2Mesh & AtlasNet &            ONet &       3D-R2N2 &   PSGN & Pix2Mesh &        AtlasNet &            ONet &            3D-R2N2 & PSGN & Pix2Mesh &        AtlasNet &            ONet \\
category    &         &      &                 &          &                 &               &        &          &                 &                 &                    &      &          &                 &                 \\
\midrule
airplane    &   0.426 &    - &           0.420 &        - &  \textbf{0.571} &         0.227 &  0.137 &    0.187 &  \textbf{0.104} &           0.147 &              0.629 &    - &    0.759 &           0.836 &  \textbf{0.840} \\
bench       &   0.373 &    - &           0.323 &        - &  \textbf{0.485} &         0.194 &  0.181 &    0.201 &  \textbf{0.138} &           0.155 &              0.678 &    - &    0.732 &           0.779 &  \textbf{0.813} \\
cabinet     &   0.667 &    - &           0.664 &        - &  \textbf{0.733} &         0.217 &  0.215 &    0.196 &           0.175 &  \textbf{0.167} &              0.782 &    - &    0.834 &           0.850 &  \textbf{0.879} \\
car         &   0.661 &    - &           0.552 &        - &  \textbf{0.737} &         0.213 &  0.169 &    0.180 &  \textbf{0.141} &           0.159 &              0.714 &    - &    0.756 &           0.836 &  \textbf{0.852} \\
chair       &   0.439 &    - &           0.396 &        - &  \textbf{0.501} &         0.270 &  0.247 &    0.265 &  \textbf{0.209} &           0.228 &              0.663 &    - &    0.746 &           0.791 &  \textbf{0.823} \\
display     &   0.440 &    - &  \textbf{0.490} &        - &           0.471 &         0.314 &  0.284 &    0.239 &  \textbf{0.198} &           0.278 &              0.720 &    - &    0.830 &  \textbf{0.858} &           0.854 \\
lamp        &   0.281 &    - &           0.323 &        - &  \textbf{0.371} &         0.778 &  0.314 &    0.308 &  \textbf{0.305} &           0.479 &              0.560 &    - &    0.666 &           0.694 &  \textbf{0.731} \\
loudspeaker &   0.611 &    - &           0.599 &        - &  \textbf{0.647} &         0.318 &  0.316 &    0.285 &  \textbf{0.245} &           0.300 &              0.711 &    - &    0.782 &           0.825 &  \textbf{0.832} \\
rifle       &   0.375 &    - &           0.402 &        - &  \textbf{0.474} &         0.183 &  0.134 &    0.164 &  \textbf{0.115} &           0.141 &              0.670 &    - &    0.718 &           0.725 &  \textbf{0.766} \\
sofa        &   0.626 &    - &           0.613 &        - &  \textbf{0.680} &         0.229 &  0.224 &    0.212 &  \textbf{0.177} &           0.194 &              0.731 &    - &    0.820 &           0.840 &  \textbf{0.863} \\
table       &   0.420 &    - &           0.395 &        - &  \textbf{0.506} &         0.239 &  0.222 &    0.218 &           0.190 &  \textbf{0.189} &              0.732 &    - &    0.784 &           0.832 &  \textbf{0.858} \\
telephone   &   0.611 &    - &           0.661 &        - &  \textbf{0.720} &         0.195 &  0.161 &    0.149 &  \textbf{0.128} &           0.140 &              0.817 &    - &    0.907 &           0.923 &  \textbf{0.935} \\
vessel      &   0.482 &    - &           0.397 &        - &  \textbf{0.530} &         0.238 &  0.188 &    0.212 &  \textbf{0.151} &           0.218 &              0.629 &    - &    0.699 &           0.756 &  \textbf{0.794} \\
\midrule
mean        &   0.493 &    - &           0.480 &        - &  \textbf{0.571} &         0.278 &  0.215 &    0.216 &  \textbf{0.175} &           0.215 &              0.695 &    - &    0.772 &           0.811 &  \textbf{0.834} \\
\bottomrule
\end{tabular}

%% file: tables/table_pcl2mesh_small.tex
\begin{tabular}{lccc}
\toprule
{} &             IoU &   Chamfer-$L_1$\footnotemark[2] & Normal Consistency \\
\midrule
3D-R2N2 &           0.565 &           0.169 &              0.719 \\
PSGN    &               - &           0.144 &                  - \\
DMC     &           0.674 &           0.117 &              0.848 \\
ONet    &  \textbf{0.778} &  \textbf{0.079} &     \textbf{0.895} \\
\bottomrule
\end{tabular}

%% file: tables/table_voxels2mesh_small.tex
\begin{tabular}{lccc}
\toprule
{} &             IoU &   Chamfer-$L_1$ & Normal Consistency \\
\midrule
Input &           0.631 &           0.136 &              0.810 \\
ONet  &  \textbf{0.703} &  \textbf{0.109} &     \textbf{0.879} \\
\bottomrule
\end{tabular}

%% file: tables/ablations_sampling.tex
\begin{tabular}{lcccc}
\toprule
{} &             IoU &   Chamfer-$L_1$ & Normal Consistency \\
\midrule
Uniform      &  \textbf{0.571} &  \textbf{0.215} &              0.834 \\
Uniform (64) &           0.554 &           0.256 &              0.829 \\
Equal        &           0.475 &           0.291 &     \textbf{0.835} \\
Surface      &           0.536 &           0.254 &              0.822 \\
\bottomrule
\end{tabular}

%% file: tables/ablations_architecture.tex
\begin{tabular}{lccc}
\toprule
{} &             IoU &   Chamfer-$L_1$ & Normal Consistency \\
\midrule
Full model &  \textbf{0.571} &  \textbf{0.215} &     \textbf{0.834} \\
No ResNet  &           0.559 &           0.243 &              0.831 \\
No CBN     &           0.522 &           0.301 &              0.806 \\
\bottomrule
\end{tabular}

%% file: parts/conclusion.tex
\section{Conclusion}
In this paper, we introduced occupancy networks, a new representation for 3D geometry.
In contrast to existing representations, occupancy networks are not constrained by the discretization of the 3D space and can hence be used to represent realistic high-resolution meshes.

Our experiments demonstrate that occupancy networks are very expressive and can be used effectively both for supervised and unsupervised learning.
We hence believe that occupancy networks are a useful tool which can be applied to a wide variety of 3D tasks.